\documentclass[10pt, a4paper]{article}
\usepackage{lrec2022} % this is the new LREC2022 Style
\usepackage{multibib}
\newcites{languageresource}{Language Resources}
\usepackage{graphicx}
\usepackage{tabularx}
\usepackage{soul}

\usepackage[hyphens]{url}
\usepackage{booktabs}
\usepackage{float}
\usepackage{multirow}
\newcommand{\lvl}{~~~}

\usepackage{subcaption}

\usepackage{titlesec}
\titleformat{\section}{\normalfont\large\bfseries\center}{\thesection.}{1em}{}
\titleformat{\subsection}{\normalfont\SmallTitleFont\bfseries\raggedright}{\thesubsection.}{1em}{}
\titleformat{\subsubsection}{\normalfont\normalsize\bfseries\raggedright}{\thesubsubsection.}{1em}{}
\renewcommand\thesection{\arabic{section}}
\renewcommand\thesubsection{\thesection.\arabic{subsection}}
\renewcommand\thesubsubsection{\thesubsection.\arabic{subsubsection}}
\usepackage{epstopdf}
\usepackage[utf8]{inputenc}

\usepackage{hyperref}
\usepackage{xstring}

\usepackage{color}

\title{Knowledge Graph -- Deep Learning: A Case Study in Question Answering in Aviation Safety Domain}

\name{Ankush Agarwal$^1$, Raj Gite$^1$, Shreya Laddha$^1$, Pushpak Bhattacharyya$^1$,  \\
{\bf \large Satyanarayan Kar$^2$, Asif Ekbal$^3$, Prabhjit Thind$^2$, Rajesh Zele$^1$, Ravi Shankar$^2$}}

\address{$^1$IIT Bombay, $^2$Honeywell, $^3$IIT Patna \\
        \{ankushagrawal, rajgite, pb\}@cse.iitb.ac.in, \{shreyaladdha, rajeshzele\}@ee.iitb.ac.in, asif@iitp.ac.in,\\
         \{satya.kar, prabhjit.thind, ravishankar.r\}@honeywell.com\\}

\abstract{
In the commercial aviation domain, there are a large number of documents, like accident reports of NTSB and ASRS, and regulatory directives ADs. There is a need for a system to efficiently access these diverse repositories to serve the demands of the aviation industry, such as maintenance, compliance, and safety. In this paper, we propose a Knowledge Graph (KG) guided Deep Learning (DL) based Question Answering (QA) system to cater to these requirements. We construct a KG from aircraft accident reports and contribute this resource to the community of researchers. The efficacy of this resource is tested and proved by the proposed QA system. Questions in Natural Language are converted into SPARQL (the interface language of the RDF graph database) queries and are answered from the KG. On the DL side, we examine two different QA models, BERT-QA and GPT3-QA, covering the two paradigms of answer formulation in QA. We evaluate our system on a set of handcrafted queries curated from the accident reports. Our hybrid KG + DL QA system, KGQA + BERT-QA, achieves 7\% and 40.3\% increase in accuracy over KGQA and BERT-QA systems respectively. Similarly, the other combined system, KGQA + GPT3-QA, achieves 29.3\% and 9.3\% increase in accuracy over KGQA and GPT3-QA systems respectively. Thus, we infer that the combination of KG and DL is better than either KG or DL individually for QA, at least in our chosen domain.
\\ \newline \Keywords{ Question Answering, Knowledge Discovery/Representation, Information Extraction, Information Retrieval} }

\begin{document}

\maketitleabstract

\section{Introduction}
An extensive set of documents is used in the aerospace sector, \textit{viz.}, system descriptions, manuals, and procedures. In addition to the industrial technical manuals, which have limited accessibility, there is a wide range of publicly available datasets related to aircraft safety. The majority of these datasets are subject to specific regulations or procedures, and safety professionals utilize them to investigate accidents and trends in aviation safety. Naturally, a Question Answering system is desired due to the large number and extensive length of such documents. For example, when investigating an accident, the ability to query the documents is required to locate similar events or find rare occurrences.
\\\\
In this paper, we present an Aviation Knowledge Graph, constructed from the accident reports, to help safety experts study the reported accidents. It contains detailed information about the accidents' cause, effect, aircraft, pilot, and meteorological conditions. KGs provides a way to integrate and relate a large number of scattered, isolated pieces of information, thus converting it into knowledge that supports making complex inferences. However, domain expertise is required to construct a KG. Besides, KGs are not as expressive as natural language sentences to explain complex events. This motivates the need for Deep Learning models such as BERT \cite{bert} from Google, and GPT-3 \cite{gpt3} from Open AI, which can fetch full-length answers from unstructured text in the documents. Our goal in this study is to combine the high coverage of textual corpora with the entity relationships in KG to improve the retrieval coverage and accuracy of the overall system.
\\\\
In this paper, we show that such an architecture relying on both Knowledge Graph and Deep Learning can benefit Question Answering. Our \textbf{major contributions} in this paper are:
\begin{enumerate}
    \item We introduce an \textbf{Aviation Knowledge Graph}, constructed from several 
    accident reports using the domain knowledge and information extraction techniques.
    \item We propose a \textbf{Question Answering system} that answers the questions asked in natural language from two modules -- (a) KG-based QA (KGQA) module, which uses a pipeline to convert Natural Language Queries to SPARQL queries and fetches responses from the Aviation Knowledge Graph, and (b) DL-based QA (DLQA) module that extracts answers from the plain text in the documents. The DLQA module has been tested using two different QA models, BERT-QA and GPT3-QA.
    
    \item We show that a \textbf{combined Question Answering system}, such as ours, outperforms the individual Knowledge Graph and Deep Learning methods on our curated test set. The combined system KGQA + BERT-QA attains \textbf{7\%} and \textbf{40.3\%} increase in accuracy over KGQA and BERT-QA modules respectively. Similarly, the other hybrid system KGQA + GPT3-QA attains \textbf{29.3\%} and \textbf{9.3\%} increase over KGQA and GPT3-QA modules respectively. Hence, we provide strong evidence that the two modules, KGQA and DLQA, complement each other, and neither of them is dispensable for the task of QA.
\end{enumerate}

This paper is organized as follows. In Section 2, we present a brief survey of the literature. Section 3 describes the public data repositories in the aviation domain relevant for building a QA system. Section 4 shows the details for constructing the Aviation Knowledge Graph and a brief explanation of its properties. Section 5 explains the overall system architecture used for Question Answering. Section 6 discusses the models, test datasets, and metrics used for evaluation. A summary of the main findings is presented in Section 7. We conclude with a direction for future work in Section 8.

\section{Related Work}
Knowledge Graph construction in the aviation field has been an important research topic. \newcite{zhao2018construction} shows the development of a KG construction system in the aviation risk field, where the American Aviation Safety Reporting System (ASRS) reports are used as an example to verify the rationality and validity of the KG construction method. \newcite{cheng2019research} built an ontology for civil aviation security by extracting knowledge from the data sources in the form of entities and relations. \newcite{wang2020research} also discusses method of extracting entities and relations from structured and unstructured text in aviation domain. However, no prior work has been done on the widely available NTSB accident reports. We exploit this gap and construct a KG from the aircraft accident reports for aviation safety.
\\\\
Querying Knowledge Graphs in natural language has been a long-standing research challenge. Early work focused on rule-based, and pattern-based systems \cite{affolter2019comparative} for the Text-to-SQL task, which later moved towards using seq2seq architecture \cite{zhong2017seq2sql} and pre-trained models \cite{he2019x} with the advent of Neural Networks (NN). We are interested in translating Natural Language Queries to SPARQL -- the standardized query language for RDF graph databases. Several QA systems have been developed that use rule-based approaches \cite{diefenbach2017wdaqua} to answer questions over DBpedia and Wikidata. Some of the prior research \cite{singh2018reinvent} divides the whole QA pipeline into smaller sub-tasks while others \cite{diefenbach2020towards} combine several components to make a pipeline. We use an approach similar to \newcite{liang2021querying} in which the translation task is divided into KG-dependent and KG-independent sub-tasks.
\\\\
Question Answering using Deep Learning has been a widely explored area in general. However, not much progress has been made in the aviation domain due to the frequently occurring %presence of 
in-domain technical jargon. A few existing works use large pre-trained transformer-based language models for Question Answering. \newcite{kierszbaum:hal-03094753} use Distilled BERT for Question Answering on ASRS reports for a small set of documents and limited test data. \newcite{airbus} employs a BM25-based retriever, followed by BERT fine-tuned for QA on a general domain data set. This model is used as a baseline for benchmarking our QA system.
\\\\
Our work is the first attempt to combine Knowledge Graph and Deep Learning in the aviation domain for Question Answering to the best of our knowledge.

% In all the prior works, either KG or DL is used individually but not as a combination in the aviation domain to perform QA.
% Knowledge Graph and Deep Learning have not been used as a combination in the aviation domain aforetime to perform QA. In all the prior works, either KG or DL is used. %all the above works, only a single concept was employed for QA. 
% In this paper, we construct a QA system %This paper introduces a Question-Answering system formed  by the synergy of KG and DL.

\section{Data Repositories}
Multiple organizations investigate aircraft accidents and warehouse all the details of casualties in a database. We identify four such public repositories, and demonstrate the use of NTSB accident reports and ADREP taxonomy in this case study. The recognized databases are the following:  

% \subsection{Accident Reports}
\begin{itemize}
    \item \textbf{Accident reports} capture all the aspects of an accident, namely aircraft specifications, pilot details, environmental state, and a comprehensive description of the suspected cause of the accident. We use two publicly available accident repositories -- \textbf{National Transportation Safety Board (NTSB) reports}\footnote{\url{https://www.ntsb.gov/Pages/AviationQuery.aspx}} (a sample report is shown in Figure \ref{ntsb} in Appendix) and \textbf{Aviation Safety Reporting System (ASRS) reports}\footnote{\url{https://asrs.arc.nasa.gov/search/reportsets.html}}. The NTSB stores investigation reports of civil aviation accidents in the US, whereas ASRS gathers information from pilots and crew members about close call events during flight journeys. These documents contain information in structured as well as unstructured format.  
    \item \textbf{Airworthiness Directives (ADs)}\footnote{\url{https://www.faa.gov/regulations\_policies/airworthiness\_directives/}} are notifications to owners of certified aircrafts about the unsafe conditions that exist in particular aircraft models and the corresponding corrective measures, which are absent in the accident reports. In our work, ADs issued by the Federal Aviation Administration (FAA) of the United States are considered. 
    \item \textbf{Accident/Incident Data Reporting (ADREP) Taxonomy}\footnote{\url{https://www.icao.int/safety/airnavigation/aig/pages/adrep-taxonomies.aspx}} (a part of the taxonomy is shown in Figure \ref{adrep} in Appendix) contains a complex multilevel hierarchy of factual descriptors (time, place, aircraft models, engine, component manufacturers, etc.) and analytical descriptors of the occurrence of accident, such as event types and explanatory factors. It is especially helpful in the construction of ontology.
\end{itemize}

\section{Aviation Knowledge Graph}
\begin{figure*}[h]
    \centering
    \includegraphics[width=1.0\linewidth]{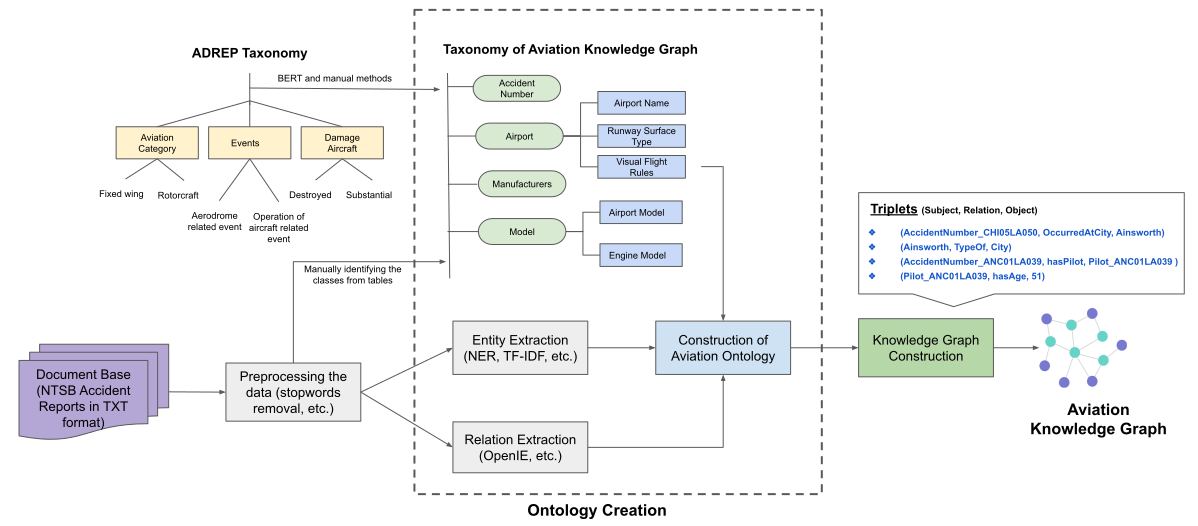}
    \caption{Aviation Knowledge Graph Construction Process from NTSB reports}
    \label{Aviation_KG}
\end{figure*}
We construct an Aviation Knowledge Graph using NTSB reports and ADREP taxonomy in Prot{\'{e}}g{\'{e}} \cite{Musen15}.
%(SPARQL query is used for retrieving answers from a Knowledge Graph). 
A total of 4000 NTSB reports from 1962 to 2015, having an average of 3000 words per report, are used in the construction. The NTSB reports consist of paragraphs (unstructured) and tables (structured) containing the information about aircraft accidents. 
% Pictures of a sample report are provided in the Appendix. 
This section discusses the pre-processing of NTSB reports, ontology creation, Knowledge Graph construction, KG evaluation, and challenges in the creation of Aviation KG. Figure \ref{Aviation_KG} shows the pipeline for building a KG from NTSB reports.

\subsection{Pre-processing}
We preprocess the NTSB reports before proceeding with entity and relation extraction. First, the NTSB reports are converted from PDF to TXT  files, followed by the application of techniques like stopword-removal, PoS tagging, and lemmatization.

\subsection{Ontology Creation}
With the help of domain experts and ADREP taxonomy, we construct an ontology from the accident reports. For example, domain knowledge from ADREP `Events' taxonomy (a snippet is shown in Figure \ref{adrep} in Appendix) is used for creating the \textit{Event} class. `Events' taxonomy is relevant because every accident report has an event sequence that gives a gist of the cause. For ontology creation, we extract the ADREP taxonomy in a tree-like data structure such that a unique path exists from the root to each leaf node. Subsequently, we obtain classes by mapping NTSB occurrences to the adequate root-to-leaf paths. We use the following two techniques for mapping:
\begin{itemize}
    \item \textbf{Mapping based on the distance between BERT embeddings}: We obtain embeddings for each path in the ADREP taxonomy and NTSB events using the BERT model. The distance is calculated between ADREP and NTSB embeddings, and the ADREP node with least distance is mapped to the NTSB event. After that, the corresponding root-to-leaf path in the ADREP tree hierarchy is associated with the NTSB event. 
    
    \item \textbf{Mapping based on Keyword Matching}: We tokenize the NTSB report events and map each token to the nodes in the ADREP taxonomy. The node with the highest similarity score is associated with the NTSB event.
\end{itemize}
Following is an example of an ADREP event mapped to NTSB occurrence (more detailed ADREP hierarchy is shown in Figure \ref{adrep}). \\
\begin{tabular}{lr}
\toprule
Sample root-to-leaf path in ADREP Taxonomy \\
\midrule
Aircraft Events      \\
\lvl Operation of the aircraft related event       \\
\lvl  Aircraft handling
related event      \\
\lvl\lvl  \textbf{Dragged wing/rotor/pod/float}    \\
\bottomrule
\end{tabular} 
\\\\
ADREP `Events' taxonomy contains a number of events and its root node is `Aircraft Events'. An event involving a `dragging of a wing' is under the category of `Aircraft handling'. This root-to-leaf path of `Dragged wing/rotor/pod/float' ADREP event is mapped to `DRAGGED WING, ROTOR, POD, FLOAT OR TAIL/SKID', which occurs as an event in an NTSB report.
\\\\
% ADREP taxonomy was also used manually, and we required observation of its use.
However, ADREP taxonomy is not always compatible with NTSB documents for creating classes. Hence, we apply following extraction techniques to the NTSB text for finding entity classes and their instances:
\begin{itemize}

\item \textbf{Named Entity Recognition (NER)} \cite{nadeau2007survey} is used to identify names, organizations, etc., from the NTSB text. These named entities are inserted as entity classes in our Ontology. For example, Long Beach, Las Vegas, and Chicago are the instances of \textit{Location} class present in NTSB reports identified by the NER method.

\item \textbf{Term Frequency -- Inverse Document Frequency (TF-IDF)} \cite{ramos2003using} technique is used to determine the important terms across NTSB documents, which are later organized as entity classes in Ontology. For example, Aircraft ID, Aircraft Damage, and Pilot Certificate are some of the important terms identified by the TF-IDF technique.

\item \textbf{C-value} \cite{frantzi2000automatic} overcomes the drawback of TF-IDF in obtaining multi-word terms. Multi-words are necessary to identify the entities such as Landing Gear, Vertical Stabilizers, etc., in aircraft reports. 

\item \textbf{Sentence Clustering} \cite{lu1978sentence} is used for grouping similar sentences and discovering the entities among them.

\item \textbf{Syntactic Analysis} technique is used where the corpus is annotated with part-of-speech tags to find hypernyms, hyponyms, and meronyms. 
We extract all the sentences containing three or more
nouns in the analysis. We then manually try to identify lexically related entities. The intuition behind selecting nouns is that most of the time, the entities (classes and instances) are tagged as nouns in the ontology. The examples of such findings are -- (a) `Scratches are found on engine’s nacelle.' From this, we can observe that \textit{nacelle} is \underline{part of} \textit{engine}.
(b) `During the engine inspection, it was observed that the wing mounted engine was working, but the fin mounted engine was not working.' Here, we observe that \textit{wing mounted engine} and \textit{fin mounted engine} are the \underline{types of} \textit{engine}.

\item \textbf{DL based techniques} are also used for entity extraction such as NER\_DL \footnote{\url{https://deeppavlov.ai/}} (trained on CoNLL dataset \cite{sang2003introduction} and uses GloVe word embeddings \cite{pennington2014glove}), Onto\_100 and NER\_DL\_BERT (trained on CoNLL dataset and uses BERT embeddings).
\end{itemize}

\textbf{Dependency Analysis} \cite{fundel2007relex} and \textbf{Open-IE} \footnote{\url{https://nlp.stanford.edu/software/openie.html}} techniques are used for relation extraction. The extracted relations are observed and inserted as properties in our Aviation Ontology. Furthermore, we manually add some properties by observing the entity classes in Ontology from NTSB reports.

\subsection{Knowledge Graph Construction}
We have an Aviation Ontology with entity classes, instances, object properties, and data properties. The entities and relations must be linked in the form of triples. We look at the entities and relations in the constructed ontology to extract triples from the NTSB reports using the regular expressions. These extracted triples are inserted into the ontology to form Knowledge Graph.
\\\\
An example for formation of triplets is as follows:
\textit{`Directional control - Not attained'} -- It is an `Aircraft Issue' present in the `Findings' section of NTSB report in tabular format (see Figure \ref{ntsb} in Appendix). In our Aviation Ontology, `Directional control' is an instance present in \textit{Aircraft\_cause} class, and `Not attained' is an instance in \textit{Aircraft\_cause\_reason} class. The triples (subject, relation, object) formed from this snippet are --
\textbf{\{Accident\_Number, isCausedByAircraftIssue, Directional control\}}, \textbf{\{Directional control, isCausedDueTo-AircraftIssue, Not attained\}}\\
\textit{Accident\_Number} is the entity class in our Aviation Ontology containing the unique accident numbers of all NTSB reports as instances. \textit{Accident\_Number} class is an essential point source for our question answering system, and thus, it is linked with \textit{Aircraft\_cause} class using relation \textit{isCausedByAircraftIssue}. In the second triplet, a relation, \textit{isCausedDueTo-AircraftIssue}, is formed for linking the \textit{Aircraft\_cause} class with \textit{Aircraft\_cause\_reason} class.
\\\\
Table \ref{app-kgproperty} describes the properties for Aviation Knowledge Graph constructed from the NTSB reports in Prot{\'{e}}g{\'{e}}. The total size of file containing the Knowledge Graph is around 12 MB.

\subsection{Knowledge Graph Evaluation}
We manually evaluate the terms obtained through entity and relation extraction techniques. A domain expert in Honeywell Corporation provided a small set of cases to validate the reach of the constructed Knowledge Graph. A total of 120 SPARQL queries with gold answers of different categories were tested, where our KG answered 83 questions, thereby achieving an accuracy of 69.1\%. 

\subsection{Challenges in Construction of Aviation Knowledge Graph}
The main challenge we faced with Aviation Knowledge Graph is scalability. As explained previously, we use many different techniques for extracting entities and relations. Later, we require manual processing for creating Ontology from these extracted entities. Similarly,  extraction of triplets from NTSB reports requires observing patterns in reports which may change with new reports. Another challenge we observed is that the domain specific terms or keywords present in NTSB reports are not identified effectively by the extraction models.

\begin{table}[H]
\begin{center}
\scalebox{0.70}{
\begin{tabularx}{\columnwidth}{|l X|}

     \hline
     \textbf{Metrics} & \textbf{Count(\#)}  \\
     \hline
     Entity Class & 239 \\
     
     Individual & 8894\\
  
     Object Property & 300 \\
   
     Data Property & 71 \\
     
     Axioms & 97879 \\
  
     Part of Relation among Classes & 494  \\
    
     Property of Relation among Classes & 353 \\
     \hline
\end{tabularx}}
\caption{Properties of Aviation Knowledge Graph: Aviation KG is constructed in Prot{\'{e}}g{\'{e}} where the class count and instances are displayed. 
}
\label{app-kgproperty}
\end{center}
\end{table}
% the extraction of entities and relations with rule-based and neural techniques because terms or keywords present in NTSB reports are related to aviation and are not identified by the extraction models.

\section{The Question Answering System}
\begin{figure}[h]
    \centering
    \includegraphics[width=\linewidth]{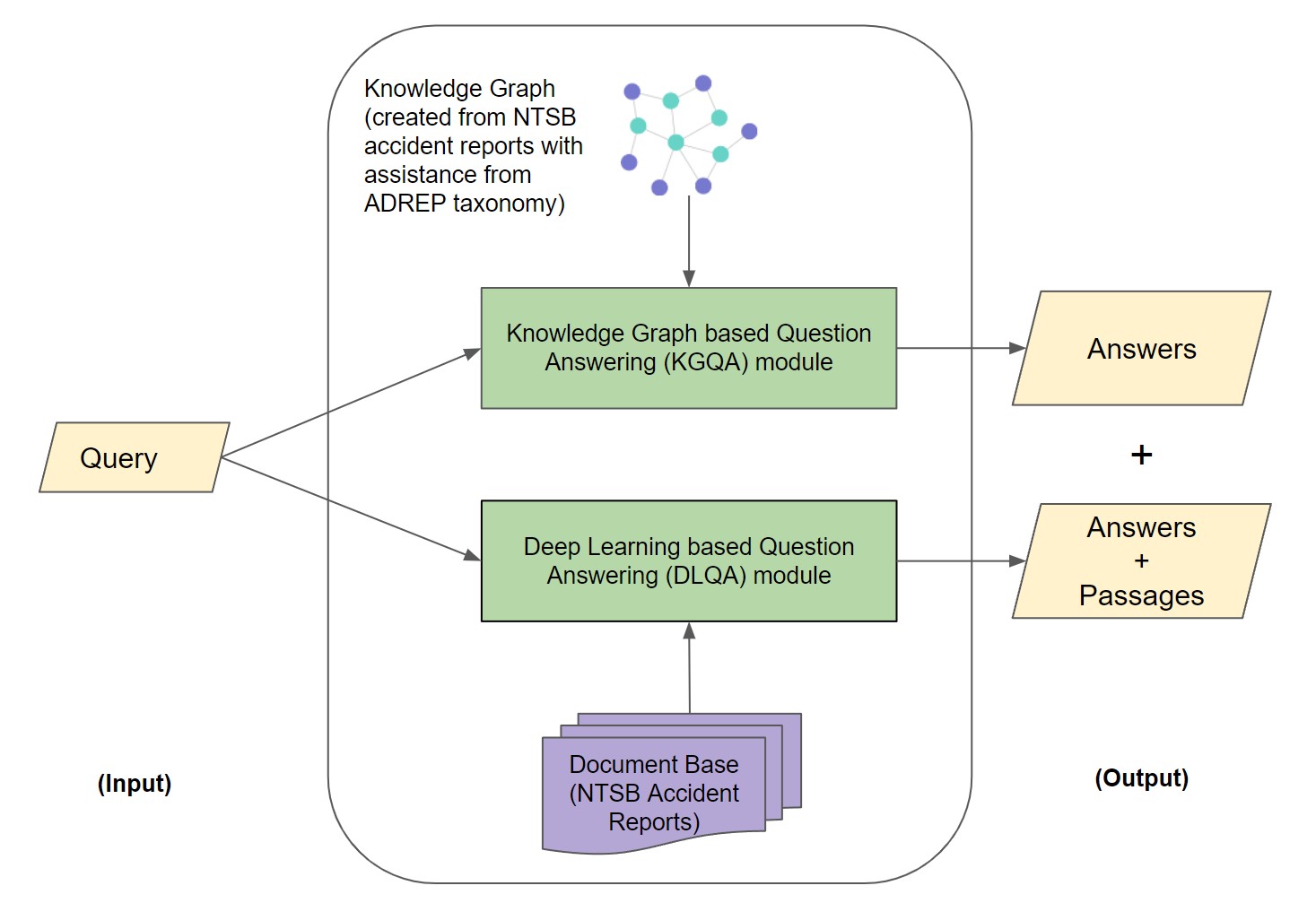}
    \caption{Knowledge Graph guided Deep Learning based Question Answering system}
    \label{QA System}
\end{figure}

Knowledge Graph guided Deep Learning based QA system is composed of two modules -- KG-based QA (KGQA) and DL-based QA (DLQA). The KGQA utilizes Aviation KG, while DLQA uses the text in the NTSB accident reports. These two modules work in parallel to answer questions as shown in Figure \ref{QA System}.

The input to the QA system is a question in natural language, which is fed to both the modules, and the output of the QA system is a combination of responses from the two modules such that the output of DLQA module follows that of the KGQA module. The output of the KGQA module is a list of answers -- entities, relations, or properties of the Aviation Knowledge Graph. DLQA module returns answers and passages whose specifications depend on the underlying model. If BERT-QA is used, we obtain a list of answer-passage pairs as response where each answer is a text span extracted from the corresponding passage. GPT3-QA returns a list of passages and an answer derived from the passage list. Figure \ref{modelqaoutput} (in Appendix) shows a sample output from KGQA and DLQA modules. In our work, we give equal importance to both answers and passages because passages help the user when answers fetched by the QA system are incorrect.
% Figure \ref{kgqaoutput} (in Appendix) shows a sample output of KGQA module. Figure \ref{bertqaoutput} (in Appendix) shows an example output of BERT-QA. 
\\\\
Our proposal of combining KG and DL is based on the following rationale:
\begin{itemize}
    \item KGQA has the advantage of domain knowledge present in the KG that captures essential details and helps in answering complex questions correctly and completely.
    \item DLQA lacks domain knowledge, but it has the advantage of coverage. With DLQA, all the text in the domain is considered for QA. This allows DLQA to answer questions that KGQA is unable to answer since KG does not capture all the information from the text.
    % With DLQA, all the texts in the domain is considered for QA, so it can answer questions about portions of text that Knowledge Graph did not cover, and thus, KGQA is unable to answer. Knowledge Graph misses some data from the text because it is tough to capture all the information into Knowledge Graph.
\end{itemize}
Section 7 provides empirical evidence that such a system formed with the synergy of Knowledge Graph and Deep Learning surpasses the performance of each individual module. In the following subsections, we discuss the design of these modules.%The below subsections discusses the internals of these modules.  

\subsection{KGQA module}

For querying Knowledge Graph stored in Prot{\'{e}}g{\'{e}}, we need to convert the Natural Language Query into a SPARQL query. We adapt an approach similar to \newcite{liang2021querying} for our NL2SPARQL pipeline (see Figure \ref{KGQA System} in Appendix) which consists of the following:
\begin{itemize}
\item \textbf{Question Type Classification}: We classify the questions into {List/Boolean/Count} categories to identify the keyword from {DISTINCT/ASK/COUNT} to use in the SELECT clause and then construct the WHERE clause in the SPARQL query.
\item \textbf{Entity-Relation Extraction}: We implement techniques such as PoS tagging, Tokenization \& Compounding, N-gram tiling, and dependency parsing for extracting entities and relations from the question. Then, we compute the similarity scores between the BERT embeddings to map the retrieved entities and relations to the corresponding entries in the underlying Knowledge Graph. This sub-task is the only KG-dependent sub-task in the pipeline.
\item \textbf{Triple Generation and Ranking}: RDF triples are constructed using all the permutations of the mapped entities and relations. However, only the valid ones are retained by performing a quick check with the Knowledge Graph by executing an ASK query. We then conduct a ranking step based on the syntactic similarity between the triples and the input question.
\item \textbf{Query Construction}: SPARQL queries are generated using the keyword specifying question type (DISTINCT for List, ASK for Boolean, COUNT for Count) and the triples obtained from the previous step. Proper PREFIX is also defined, specifying the base Knowledge Graph used.
\end{itemize}
The constructed SPARQL queries are executed over the Aviation Knowledge Graph to retrieve answers from KG which are either subjects or objects in the triples. The KGQA does not produce any response when the pipeline cannot form a valid triple in the Triple Generation \& Ranking sub-task.

\subsection{DLQA module}
We examine two models, namely BERT-QA and GPT3-QA \cite{gpt3}, which form two independent DLQA modules. The reason for reviewing BERT-QA and GPT3-QA is to cover the two paradigms of answer formulation in QA, namely extractive and abstractive. Both the models require a collection of passages from the documents. BERT-QA extracts a text span from the relevant passage as an answer (extractive QA), while GPT3-QA generates text from the relevant passages (abstractive QA). Both these models are based on the retriever-reader pipeline as follows:

\begin{itemize}
    \item \textbf{Retriever}: The retriever retrieves top-k relevant passages given a question. It projects the question and all the passages in the collection to a semantic vector space, such that passages relevant to a question are positioned closer to the question as compared to less relevant passages. This projection of query and passages into semantic vector space is accomplished by Sentence-BERT \cite{sbert} in case of BERT-QA and by GPT3 ‘ada’ model in case of GPT3-QA \footnote{\url{https://beta.openai.com/docs/guides/answers}}. Passages relevant to the question obtain a higher relevance score, and top-k passages are passed to the reader.
    
    \item \textbf{Reader}: The reader extracts/generates answers from the relevant passages given a question. It is a Machine Reading Comprehension (MRC) task where the underlying model understands the context passage and tries to answer the given question. In BERT-QA, we employ BERT \cite{bert} fine-tuned on MRC task as the reader, which extracts text spans from the passages as the answer. In GPT3-QA, we utilize GPT3 ‘curie’ model as the reader, which generates text from the passages as an answer.

\end{itemize}

Figure \ref{ret-read} (in Appendix) shows the retriever-reader pipeline along with an example.
%Need to finish the section(QA system) with some conclusion or any statement.

\section{Experimental Setup}
%Clearly not understandable what is in experimental section. 1-2 lines intro will help.

\textbf{Data Pre-processing: }The NTSB reports, available as PDFs, are converted to TXT format for constructing the Knowledge Graph (as mentioned in Section 4). 
% Next, the reports are converted to specific format files used by the DL models. 
Passages from these reports are extracted and stored as JSON objects with three properties -- passage heading, passage text, and report ID. The resulting JSON files are used by the BERT-QA model. The reports are also converted to JSONL format, where each row contains the extracted passages. These JSONL files are consumed by the GPT3-QA model. 
\\\\
\textbf{Models:} The details of the models used in our Question Answering system are as follows:
\begin{itemize}
    \item The \texttt{bert-base-nli-mean-tokens} \footnote{\url{https://huggingface.co/sentence-transformers/bert-base-nli-mean-tokens}} model from the Sentence Transformers  \cite{reimers-2019-sentence-bert} library is used for similarity calculation in the \textbf{KGQA} module.
    
    \item In the \textbf{BERT-QA} model, we use a fine-tuned Sentence BERT model, namely \texttt{multi-qa-MiniLM-L6-cos-v1}\footnote{\url{https://huggingface.co/sentence-transformers/multi-qa-MiniLM-L6-cos-v1}} from the Sentence Transformers library, as the retriever. This Sentence BERT model is fine-tuned for the Semantic Search task using 215M question-answer pairs. Similarly, a BERT model, \texttt{deepset/bert-base-cased-squad2}\footnote{\url{https://huggingface.co/deepset/bert-base-cased-squad2}} from the Transformers library \cite{wolf-etal-2020-transformers}, fine-tuned on MRC task using SQuAD 2.0 dataset \cite{rajpurkar2018know} is used as the reader.
    
    \item A \textbf{baseline} QA system is built on a similar approach as BERT-QA, called \textbf{BM25-BERT}. In this model, BM25 \cite{bm25} relevance score is used for retrieving most relevant passages using sparse lexical features like TF-IDF.  The BERT model in BM25-BERT has the same configurations as the reader in the BERT-QA model.
%     Not clear as to what the baseline is. Last line is dicey. There are two BERT models in BERT-QA - so added 'as the reader in the'
    \item From various flavors of \textbf{GPT-3}, we chose a combination of \textit{ada} and \textit{curie} as retriever and reader respectively, in the GPT3-QA model, based on offline evaluations.
% add 'in the GPT3-QA model'??
\end{itemize}
% Our QA system, which comprises KGQA and DLQA modules, is configured to output ten results such that DLQA results follow KGQA results. When both the modules have adequate results to display, each module displays a maximum of five results, and when KGQA doesn’t have enough results, DLQA adjusts the result count to ten by contributing more results.  
 Our QA system, which comprises of KGQA and DLQA modules, is configured to output ten responses such that the outputs from the DLQA follow those from KGQA. When both the modules have sufficient outputs to display, each module returns its top five results, else DLQA adjusts its return count to make up for the shortfall in the KGQA responses.
% why dlqa outputs more?
\\\\
\textbf{Test Dataset:} For evaluating the performance of our QA system, a test set is curated from 50 NTSB reports. We restrict the number to 50 by considering the difficulty in incorporating all 50 reports while creating a single test instance. All these reports belong to the year 2002. A total of 150 test instances were created (a single instance of test set is shown in Figure  \ref{app-testset} in Appendix), where each instance consists of a query, a list of actual answers, and a list of passages where the answers can potentially be found. The entire test set is manually curated by looking up every query in those 50 reports and recording the desired answers and the paragraphs. We tag each paragraph with the unique accident number in its report, which aids in evaluating the retriever sub-module of DLQA. We do not evaluate our QA system with standard QA datasets because the KG used in KGQA module is domain-specific, and thus, does not contribute when our QA system operates on other domains.  
\\\\
\textbf{Evaluation Metrics:} Our overall system outputs both answers and passages. Sometimes, the predicted answer may not be correct, so reading the passage allows the user to reasonably satisfy the query intent. Thus, both the accuracy of answers and passages become important, and we evaluate both separately. Finally, we report the performance using four metrics: 
\begin{itemize}
    \item \textbf{Exact Match} (binary value): An exact match occurs if the first answer predicted by the system is present in the list of actual answers exactly. 
    \item \textbf{Exact Recall}: We determine exact recall as the fraction of actual answers available exactly in the top 10 answers predicted by the system. However, for more than 10 actual answers, we consider only a subset such that they contain the maximum number of predicted answers. This is done to ensure a recall $\leq 1$ since we predict 10 answers only. 
    % However, for more than 10 actual answers, we keep the denominator as 10 since we only take into account 10 predicted answers. 
    \item \textbf{Semantic Accuracy} (binary value): We use this as a flexible alternative to Exact Match, which relies solely on lexical overlap. We define our output as semantically accurate if any value in the list of top 10 predicted targets is semantically similar to any value in the list of actual targets. This metric does not consider the rank of the predicted targets, unlike Exact Match.
    \item \textbf{Semantic Recall}: We compute this metric as the fraction of actual targets which are semantically similar to the top 10 predictions of the system. Similar to Exact Recall, we consider only a subset of 10 targets such that they contain the maximum number of predicted targets.
\end{itemize}
% Average
Exact Match and Exact Recall are calculated only for the answers predicted by the system, whereas Semantic Accuracy and Semantic Recall are computed for both the answers and the passages. Exact Match and Semantic Accuracy are a measure of correctness of our Question Answering system, whereas Exact Recall and Semantic Recall estimate the completeness of the outputs of our system. 

\section{Results and Analysis}

\begin{table*}[!h]

\begin{tabularx} {\columnwidth}{lccrrrr}

     \cline{1-7}
\multicolumn{1}{c}{\textbf{ \centering Model}} & \multicolumn{2}{c}{Answers} & \multicolumn{2}{c}{Answers} &
\multicolumn{2}{c}{Passages}\\
\cmidrule(lr){2-3} \cmidrule(lr){4-5} \cmidrule(lr){6-7}
&
\multicolumn{1}{p{1.6cm}}{\centering Exact\\ Match} & \multicolumn{1}{p{1.6cm}}{\centering Exact\\ Recall} & \multicolumn{1}{p{1.6cm}}{\centering Semantic \\ Accuracy} & \multicolumn{1}{p{1.6cm}}{\centering Semantic \\ Recall} & \multicolumn{1}{p{1.6cm}}{\centering Semantic \\ Accuracy} & \multicolumn{1}{p{1.6cm}}{\centering Semantic \\ Recall} \\ 
     \cline{1-7}
      BM25-BERT (baseline) & \multicolumn{1}{p{1.6cm}}{\centering0.013} & \multicolumn{1}{p{1.6cm}}{\centering 0.153} & \multicolumn{1}{p{1.6cm}}{\centering 0.113} & \multicolumn{1}{p{1.6cm}}{\centering 0.143} & \multicolumn{1}{p{1.6cm}}{\centering 0.333} & \multicolumn{1}{p{1.6cm}}{\centering 0.253} \\
      \cline{1-7}
   
KGQA & \multicolumn{1}{p{1.6cm}}{\centering 0.347} & \multicolumn{1}{p{1.6cm}}{\centering 0.345} & \multicolumn{1}{p{1.6cm}}{\centering 0.560} & \multicolumn{1}{p{1.6cm}}{\centering 0.545} & \multicolumn{1}{p{1.6cm}}{\centering -} & \multicolumn{1}{p{1.6cm}}{\centering -}  \\
    
BERT-QA & \multicolumn{1}{p{1.6cm}}{\centering 0.040} & \multicolumn{1}{p{1.6cm}}{\centering 0.157} & \multicolumn{1}{p{1.6cm}}{\centering 0.227} & \multicolumn{1}{p{1.6cm}}{\centering 0.217} & \multicolumn{1}{p{1.6cm}}{\centering 0.393} & \multicolumn{1}{p{1.6cm}}{\centering 0.311} \\
  
GPT3-QA & \multicolumn{1}{p{1.6cm}}{\centering \textbf{0.600}} & \multicolumn{1}{p{1.6cm}}{\centering 0.547} & \multicolumn{1}{p{1.6cm}}{\centering 0.760} & \multicolumn{1}{p{1.6cm}}{\centering 0.620} & \multicolumn{1}{p{1.6cm}}{\centering 0.813} & \multicolumn{1}{p{1.6cm}}{\centering 0.734}\\
  \cline{1-7}
KGQA + BERT-QA & \multicolumn{1}{p{1.6cm}}{\centering 0.349} & \multicolumn{1}{p{1.6cm}}{\centering 0.403} & \multicolumn{1}{p{1.6cm}}{\centering 0.630} & \multicolumn{1}{p{1.6cm}}{\centering 0.588} & \multicolumn{1}{p{1.6cm}}{\centering 0.788} & \multicolumn{1}{p{1.6cm}}{\centering 0.715}\\
     
\textbf{KGQA + GPT3-QA} & \multicolumn{1}{p{1.6cm}}{\centering 0.500} & \multicolumn{1}{p{1.6cm}}{\centering \textbf{0.628}} & \multicolumn{1}{p{1.6cm}}{\centering \textbf{0.853}} & \multicolumn{1}{p{1.6cm}}{\centering \textbf{0.680}} & \multicolumn{1}{p{1.6cm}}{\centering \textbf{0.866}} & \multicolumn{1}{p{1.6cm}}{\centering \textbf{0.784}}
      \\
     \cline{1-7}
\end{tabularx}
\caption{Evaluation results of `Answers' and `Passages' predicted by Question Answering models on Exact Match, Exact Recall, Semantic Accuracy and Semantic Recall metrics. In most metrics, KGQA + GPT3-QA performs better compared to other models.}
\label{table1}
\end{table*}

This section analyzes the performance of 6 models -- BM25-BERT (baseline), KGQA, BERT-QA, GPT3-QA, KGQA + BERT-QA, and KGQA + GPT3-QA. The last two models are the combined models that use both KG and DL, whereas the first four are individual models. We evaluate both the answers and passages predicted by these models. Table \ref{table1} presents the evaluation results of answers on Exact Match and Exact Recall metrics as well as the evaluation results of answers and passages on Semantic Accuracy and Semantic Recall metrics. As the KGQA system only outputs answers, we consider the answers from KGQA as passages when evaluating passages in KGQA + BERT-QA and KGQA + GPT3-QA models. 
\\\\
It is trivial to see that the Exact Match metric scores are relatively lower than the Semantic Accuracy metric scores because of the dependence of Exact Match on only the lexical aspect of answers. Thus, we are more interested in Semantic Accuracy and Semantic Recall values. Each of the KGQA, BERT-QA, and GPT3-QA models beat the baseline model on every metric. KGQA system performs better than the BERT-QA model, which validates the importance of domain knowledge. GPT3-QA model is the best performing among the individual models owing to its large size and extensive pre-training.
\\\\
Our combined models perform better than their counterparts in most cases. Since the top answers from each system are merged to generate top-10 answers for the combined KG + DL system, accuracy and recall scores increase. However, due to the fact that our system outputs KGQA answers at the top of DLQA, the Exact Match score of KGQA + GPT3-QA is lower than that of GPT3-QA. In other scenarios, the KGQA + GPT3-QA system achieves an increase of 29.3\% and 9.3\% increase in answer accuracy over KGQA and GPT3-QA respectively. The KGQA + BERT-QA system also attains an increase in answer accuracy of 7\% over KGQA and 40.3\% over BERT-QA. A similar trend is seen in passage retrieval accuracy for both the hybrid QA models.
% The KGQA + BERT-QA system achieves a 7\% increase in accuracy for predicted answers over KGQA and a 47.3\% increase in accuracy of passage retrieval over BERT-QA. 
Overall, \textbf{KGQA + GPT3-QA is the best model for the QA task}.
\\\\
% \textit{Analysis}\\
\textbf{Limitations of KGQA module}: The low accuracy of the KGQA model can be attributed to the following factors -- (a) Natural language interface -- KGQA can answer questions efficiently only when valid SPARQL queries can be formed. The formation of the SPARQL queries relies heavily on the NL2SPARQL pipeline, which being rule-based, is prone to errors. (b) Coverage -- Not every sentence can be converted to a triple format; hence the amount of information present in the Knowledge Graph is limited. For example, the question \textit{`What discrepancy was noted due to which flight landed at La Belle Municipal Airport?'}, which expects an answer `problem with fuel gauge' cannot be answered using our KG as there exists no valid triple in the KG containing such information.
\\\\
\textbf{Limitations of DLQA module}: The BERT-QA system suffers from the lack of fine-tuning on the domain dataset and hence the scores for this model are very close to the baseline. We notice that GPT3-QA does not answer multi-hop questions adequately, \textit{i.e.}, questions that can only be answered based on the information in two different paragraphs of the same document or multiple documents. For example, the question \textit{`Which accidents involved aircraft operated by Johnny Thornley and manufactured by Subaru?'} is not answered correctly by GPT3-QA. One needs to look at two different paragraphs in a document for fetching answers, where GPT3-QA fails. However, the responses to such questions can be obtained by traversing multiple triples in the Knowledge Graph.
\\\\
\textbf{Strengths of the combined system (KGQA + DLQA)}: Our combined system overcomes the shortcomings of the individual systems. It answers both the questions listed in the above paragraphs correctly. This combination solves the coverage issue of Knowledge Graph and the lack of domain knowledge in Deep Learning. The questions which remain unanswered by the KGQA module are answered by the DLQA module and vice versa. Both the modules complement each other, and thus the system formed by their synergy achieves better accuracy than the individual components.
\\\\
The resources contributed by us are publicly available on GitHub\footnote{\url{https://github.com/RajGite/KG-assisted-DL-based-QA-in-Aviation-Domain}}. 

\section{Conclusion and Future Work}
%KGQA works better than simple pre-trained models- domain knowledge.
Aircraft safety is indispensable in the aviation domain. Aircraft accidents are unfortunate, and thus the reported accidents must be studied carefully. We have successfully created an Aviation Knowledge Graph from NTSB aircraft accident reports to help experts in their study. We also provide a Knowledge Graph guided Deep Learning based Question Answering system, which outperforms the individual KGQA and DLQA systems. The dominance of the combined QA system is proved theoretically and empirically by evaluating it on our handcrafted test set.
\\\\
Currently, the constructed Knowledge Graph is based on the NTSB reports. We will continue our work on Knowledge Graph construction for ASRS reports and ADs. We aim to merge all the KGs to expand our knowledge scope in aviation safety. KGs can be extensively utilized if there is a robust querying mechanism. Thus, we need to improve our NL2SPARQL mechanism so that complex Natural Language Queries can be converted correctly to SPARQL queries. We have shown that KGQA + GPT3-QA model performs best on most metrics. We plan to improve the system by devising a mechanism to re-rank the results of the combined system or by implementing a solid combination framework.

%\section{Acknowledgements}
%This research is supported by Government of India. We thank Amit Patil for his contribution in development of Aviation Knowledge Graph. 

\nocite{*}
\section{Bibliographical References}\label{reference}
%\label{main:ref}

\bibliographystyle{lrec2022-bib}
\bibliography{lrec2022-example}

% \section{Language Resource References}
% \label{lr:ref}
% \bibliographystylelanguageresource{lrec2022-bib}
% \bibliographylanguageresource{languageresource}

\section{Appendix}

% \subsection{Data Repositories utilized for constructing Aviation KG}\label{app-datarepo}

\begin{figure}[H]
    \centering
    \includegraphics[width=0.9\linewidth]{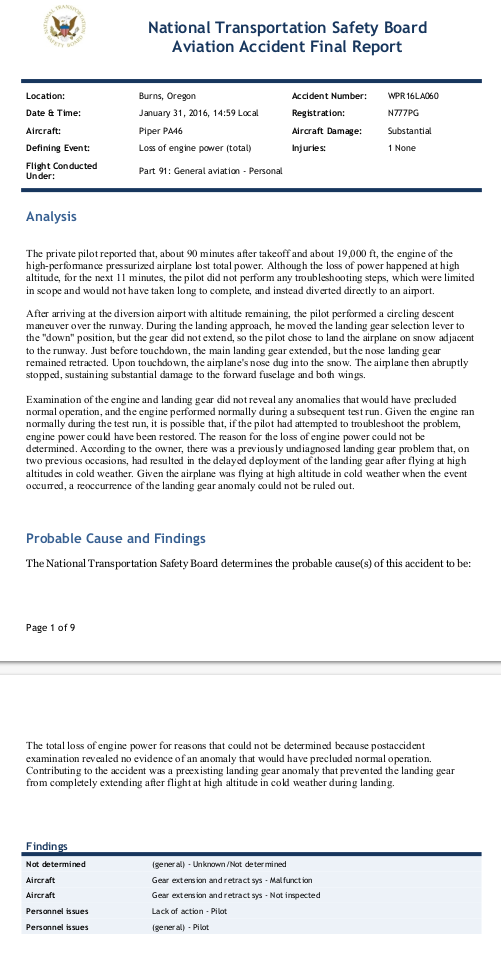}
    \caption{A snippet of NTSB report }
    \label{ntsb}
\end{figure}

\begin{figure}[H]
    \centering
    \includegraphics[width=0.9\linewidth]{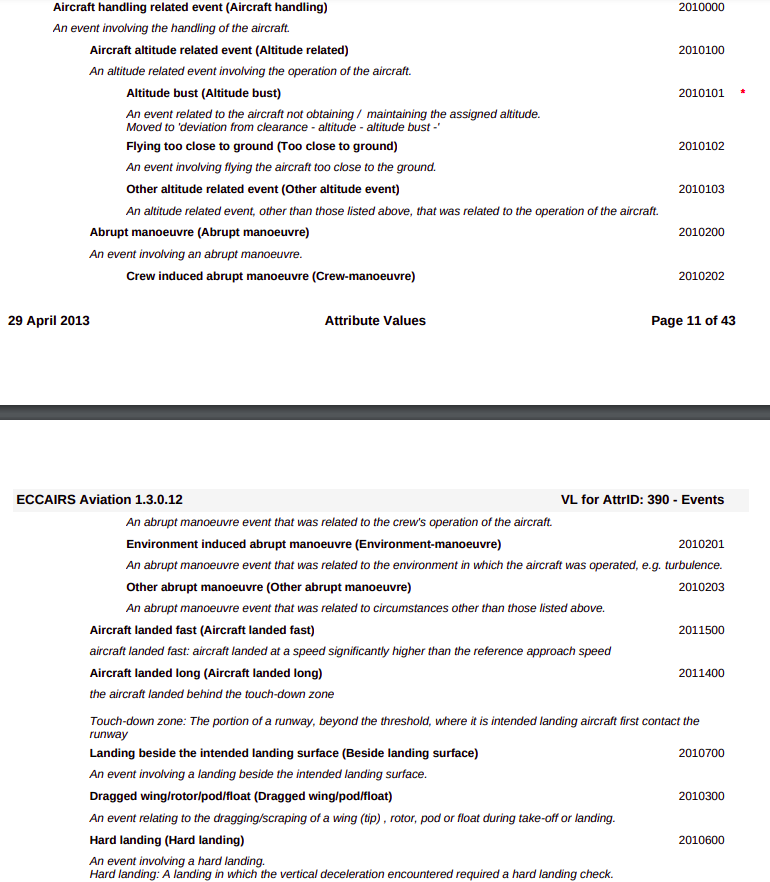}
    \caption{A snippet of ADREP Events Taxonomy}
    \label{adrep}
\end{figure}

% \subsection{Properties of Aviation KG}\label{app-kgproperty}

% \subsection{Examples defining the QA task}\label{app-qa-eg}
\begin{figure}[h]
\begin{subfigure}[b]{0.5\textwidth}
    \centering
    \includegraphics[width=0.68\linewidth]{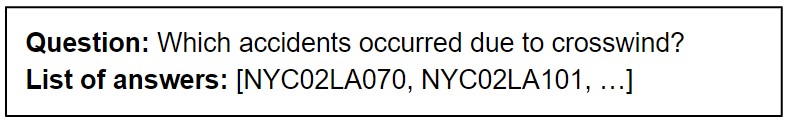}
    \caption{An example depicting the output of KGQA module}
    \label{kgqaoutput}
\end{subfigure}
\begin{subfigure}[b]{0.5\textwidth}
    \centering
    \includegraphics[width=0.68\linewidth]{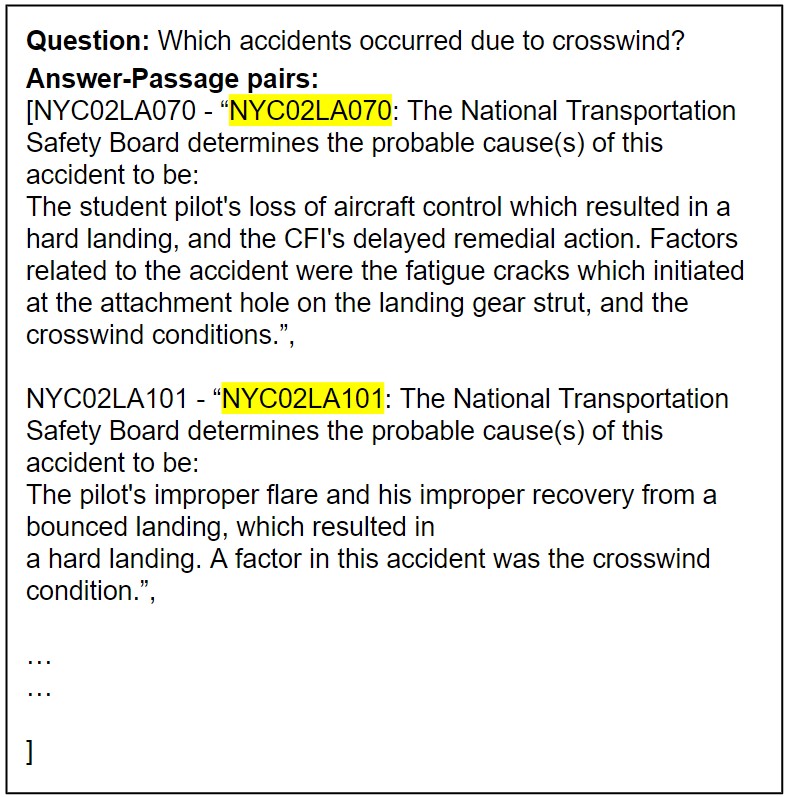}
    \caption{An example depicting the output of DLQA module when the underlying model is BERT-QA}
    \label{bertqaoutput}
\end{subfigure}
\begin{subfigure}[b]{0.5\textwidth}
    \centering
    \includegraphics[width=0.68\linewidth]{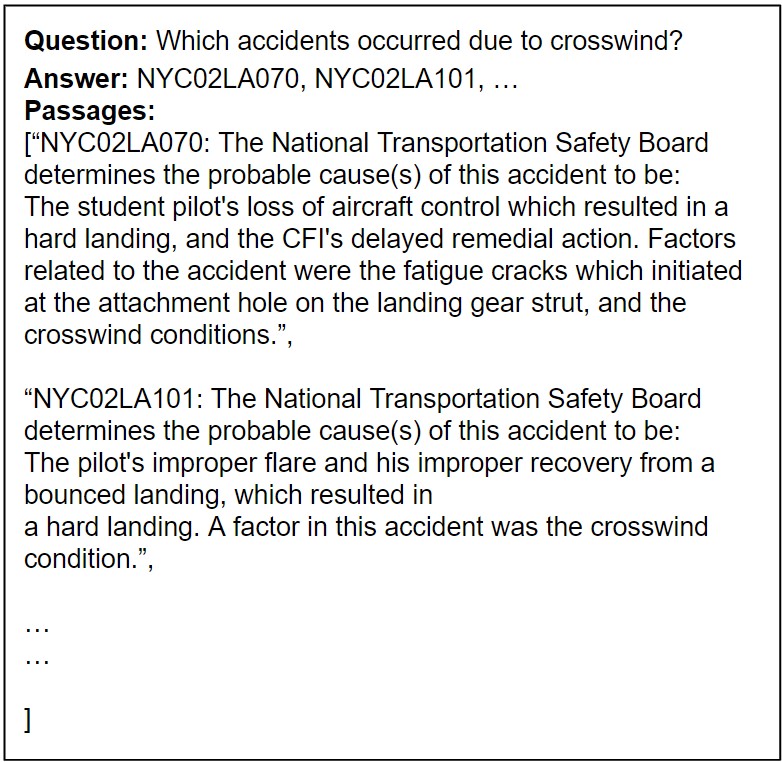}
    \caption{An example depicting the output of DLQA module when the underlying model is GPT3-QA}
    \label{gpt3qaoutput}
\end{subfigure}
\caption{Examples defining QA task}
\label{modelqaoutput}
\end{figure}

% \subsection*{Test Dataset for QA}\label{app-testset}
% Below is a single instance in our Test Dataset:\newline
% [\{\textbf{Question} : Which accidents are caused due to pilot's failure to maintain control leading to collision?\}, \\
% \{\textbf{Answers} : SEA02FA036 , FTW02LA170 \}, \\
% \{\textbf{Passages} : \{\textit{SEA02FA036}: The pilot's failure to maintain aircraft control during takeoff.\}, \{\textit{FTW02LA170}: Probable Cause and Findings: The National Transportation Safety Board determines the probable cause(s) of this accident to be: the pilot's failure maintain directional control of the airplane during a touch and go landing which resulted in a collision with trees. A contributing factor was the gusting wind conditions\}]

\begin{figure}[h]
    \centering
    \includegraphics[width=0.65\linewidth]{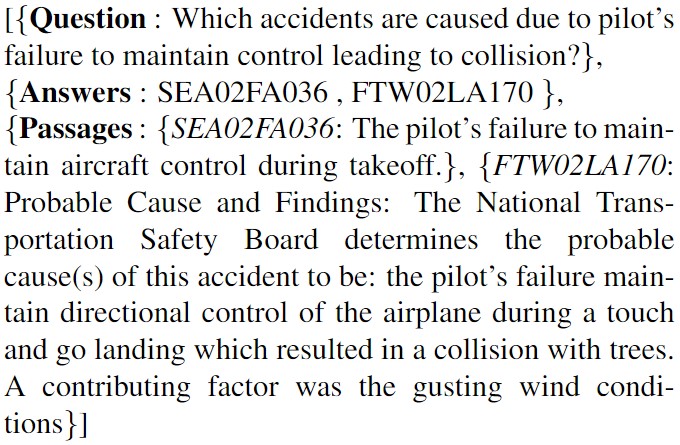}
    \caption{A single instance in the QA Test Set}
    \label{app-testset}
\end{figure}

% \subsection{Pipelines involved in the QA system}\label{app-qa-pipeline}
\begin{figure*}[t]
    \centering
    \includegraphics[width=\linewidth]{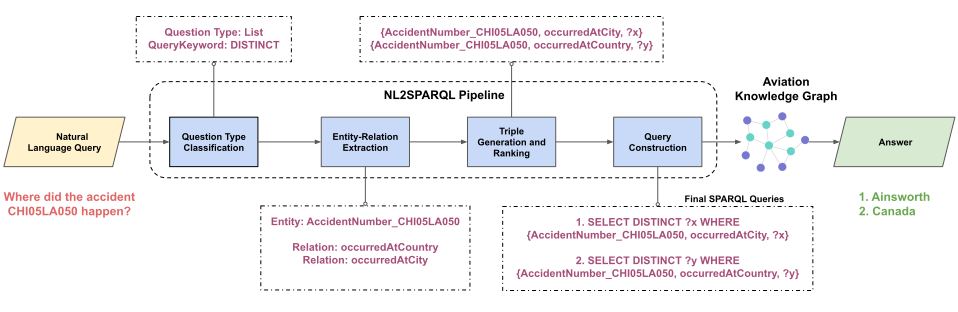}
    \caption{NL2SPARQL pipeline for KGQA system}
    \label{KGQA System}
\end{figure*}

\begin{figure*}[h]
    \centering
    \includegraphics[width=\linewidth]{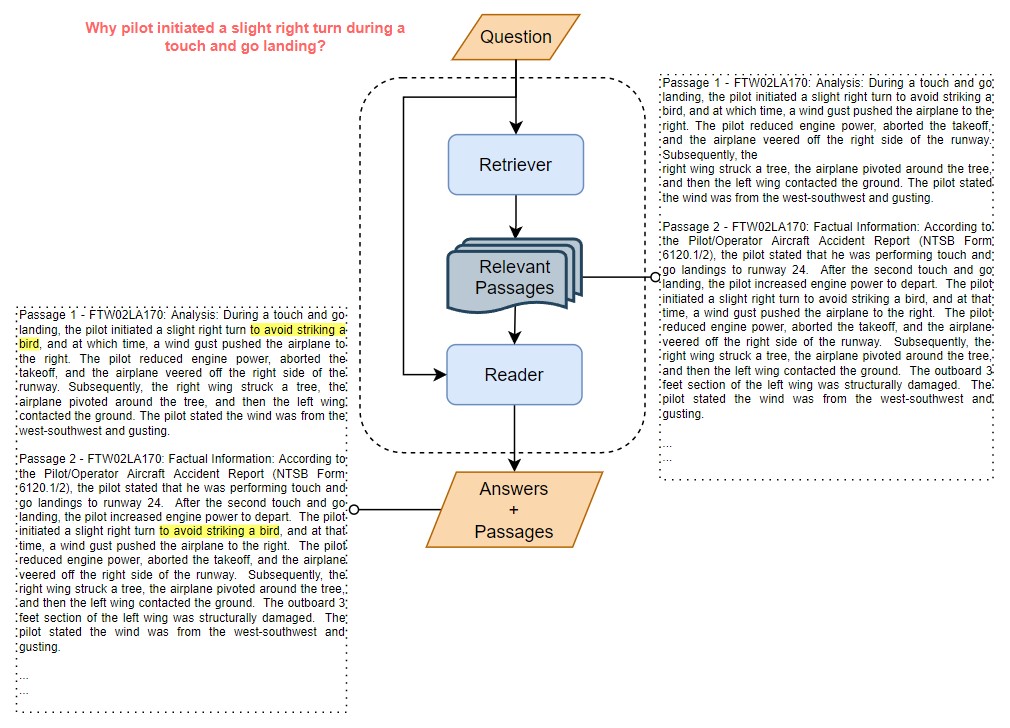}
    \caption{Retriever-Reader pipeline for DLQA system. The highlighted text in the passage is the answer extracted by the DLQA system.}
    \label{ret-read}
\end{figure*}

\end{document}